\pdfoutput=1
\documentclass[11pt]{article}
\usepackage{emnlp2021}
\usepackage{times}
\usepackage{latexsym}
\usepackage{multirow}
\usepackage{graphicx}
\usepackage{amsmath}
\usepackage{amssymb}
\usepackage{booktabs}
\usepackage{subcaption}
\usepackage{caption}

\newcommand{\br}{\mathbf{r}}

\newcommand{\ie}{\textit{i.e.}}
\newcommand{\eg}{\textit{e.g.}}

\usepackage[T1]{fontenc}
\usepackage[utf8]{inputenc}
\usepackage{microtype}
\title{MCP: Self-supervised Pre-training for Personalized Chatbots with Multi-level Contrastive Sampling}
\author{Zhaoheng Huang$^1$, Zhicheng Dou$^{1}$\thanks{$^*$Corresponding author.} , Yutao Zhu$^2$, \and Zhengyi Ma$^1$ \\
$^1$Gaoling School of Artificial Intelligence, Renmin University of China, Beijing, China \\
$^2$University of Montreal, Quebec, Canada \\
\texttt{\{zhaoheng\_huang,dou,zymaa\}@ruc.edu.cn, yutao.zhu@umontreal.ca}
}
\begin{document}
\maketitle
\begin{abstract}
Personalized chatbots focus on endowing the chatbots with a consistent personality to behave like real users and further act as personal assistants.
Previous studies have explored generating implicit user profiles from the user's dialogue history for building personalized chatbots.
However, these studies only use the response generation loss to train the entire model, thus it is prone to suffer from the problem of data sparsity.
Besides, they overemphasize the final generated response's quality while ignoring the correlations and fusions between the user's dialogue history, leading to rough data representations and performance degradation.
To tackle these problems, we propose a self-supervised learning framework MCP for capturing better representations from users' dialogue history for personalized chatbots.  
Specifically, we apply contrastive sampling methods to leverage the supervised signals hidden in user dialog history, and generate the pre-training samples for enhancing the model.
We design three pre-training tasks based on three types of contrastive pairs from user dialogue history, namely response pairs, sequence augmentation pairs, and user pairs.
We pre-train the utterance encoder and the history encoder towards the contrastive objectives and use these pre-trained encoders for generating user profiles while personalized response generation.
Experimental results on two real-world datasets show a significant improvement in our proposed model MCP compared with the existing methods.
\end{abstract} 

\section{Introduction}
In recent years, open-domain chatbots have achieved impressive progress in the natural language processing~(NLP) field~\cite{DBLP:conf/acl/LiGBSGD16,DBLP:conf/aaai/ZhouHZZL18,DBLP:conf/emnlp/ChanLYCHZY19,DBLP:conf/emnlp/LiuQXW20,DBLP:conf/ecir/ZhuNZDD21}. 
Given the input utterance, the chatbot responds with an appropriate response. However, for the same input utterance, chatbots usually provide similar responses for all users. 
This one-size-fits-all strategy cannot satisfy the varying needs of users and is lean to generate safe but meaningless responses, such as ``I don't know''~\cite{DBLP:conf/naacl/LiGBGD16,DBLP:journals/ir/ZhuDNW20}. 
To solve this problem, some researchers have begun to endow chatbots with personality and develop personalized chatbots~\cite{Zhang_2018_dog,DBLP:conf/sigir/madousigir21,DBLP:conf/naacl/ZhongD0QW22}. 
When equipped with personal information (either given by a predefined profile or learning from dialogue history), personalized chatbots can generate more user-specific and informative responses.

Although the existing methods for personalized chatbots are capable of generating some personalized responses, two major shortcomings hurt user profile building and response generation.
\textbf{First}, like non-personalized chatbots, they only use the single response generation loss to train the entire model.
It has been found that such an optimization way is easy to suffer from the data sparsity scenarios, where users may have limited dialogue history in real-world applications~\cite{DBLP:conf/cikm/SongS0DX0T19}.
\textbf{Second}, they overemphasize the quality of the final generated response, while the correlations and the fusion between the user's historical responses are not captured in the data representations. 
Such rough data representations are used for building inaccurate user profiles and hurt the generalizability of the model. 
Recent studies in pre-trained language models have shown that effective data representation is a key factor to improve the performance of various downstream tasks~\cite{DBLP:conf/iclr/KongdYLDY20}. 
Therefore, there is a need to rethink the learning paradigm for developing more effective personalized chatbots.

To address the above issues, we propose to \textbf{apply the self-supervised learning framework into personalized chatbots}. 
We design three novel pre-training tasks to leverage the correlations and self-supervised signals brought by dialogue history. 
Self-supervised learning has achieved great success in various NLP~\cite{DBLP:conf/naacl/DevlinCLT19, DBLP:conf/emnlp/BansalGWMM21} and Information Retrieval~\cite{DBLP:conf/sigir/GuoMMQZJCD22,DBLP:conf/cikm/MaDXZJCW21,DBLP:conf/cikm/0001NDMZDZJ21,DBLP:conf/cikm/ZhouD0W21} tasks.
It aims to let the model learn the intrinsic structure of the raw data without additional supervised signals. 
In fact, the user dialogue history contains massive supervised signals for capturing personalized information. 
For example, if one user gave two responses to the same user within a short time, these two responses tend to be semantically relevant, thus the model should build similar representations for these two responses. 
Such supervised signals hidden in dialogue history can help the model to build more accurate data representations, and we can construct amounts of training samples for pre-training the model in a self-supervised manner. 
It can be seen that the self-supervised learning framework can perfectly help to solve two major issues: rough data representations and data sparsity.

Based on the above observations, we propose a novel self-supervised learning approach \textbf{MCP} for personalized chatbots that is a two-stage neural framework and designs three pre-training tasks based on the dialogue history. 
To build the user profile from the user's dialogue history, we design two encoders for learning the single utterance representation and the user profile representation, \ie, \textbf{utterance encoder} and \textbf{history encoder}. 
The utterance encoder aims to learn the semantics and representations of user responses, and the history encoder focuses on encoding the user's historical response sequence and building the user profile representation.
We design a two-stage framework for applying self-supervised learning to personalized chatbots. 
\textbf{In the first stage}, we design three pre-training tasks to generate supervised signals from the dialogue history and pre-train the utterance encoder and the history encoder towards these designed objectives. 
We construct three types of contrastive pairs from the user's dialogue history as the pre-training samples, namely response pairs, sequence augmentation pairs, and user pairs. 
Via such a pre-trained method, our model can learn better data representations based on the pre-training data, further adapting data representations to personalized chatbot scenarios. 
\textbf{In the second stage}, we use the parameters of the pre-trained encoders to initialize the encoders in the personalized encoder-decoder model and use the utterance encoder and history encoder to build the user's profile representation. 
The user profiles are fed into the encoder-decoder model to drive the personalized response generation process. 
To verify the effectiveness of our model, we conduct extensive experiments on two large-scale datasets, \ie, Weibo and Reddit. 
Experimental results show that MCP achieves state-of-the-art performance compared to a number of competitive methods. 

Our contributions can be summarized as follows: 
(1) We propose a two-stage framework for applying self-supervised learning to train personalized chatbots for better data representation.
To the best of our knowledge, it is the first time that pre-training is leveraged based on a user's dialogue history for personalized chatbots.
(2) We design three pre-training tasks based on the correlations and supervised signals hidden in the user's dialogue history. We construct response pairs, sequence augmentation pairs, and user pairs from different views of the dialogue history for better pre-training the encoder models.
(3) The experimental results on two real-world large-scale datasets show the effectiveness of our proposed model.

\section{Related Work}\label{related}
\noindent\textbf{Open-domain Generation-based Chatbots.}
Here we briefly introduce some generation-based methods, as they are most relevant to ours.
In some early studies, dialogue response generation was considered a statistical machine translation problem ~\cite{DBLP:conf/emnlp/RitterCD11}.
Nowadays, sequence-to-sequence (Seq2Seq) models are the mainstream method due to the fast development of deep learning and neural network~\cite{DBLP:conf/aaai/SerbanSBCP16}. 
Many studies have tried to endow Seq2Seq-based methods with human attributes, such as:
(1) introducing additional commonsense knowledge to generate a more reasonable response~\cite{DBLP:conf/ijcai/ZhouYHZXZ18}; 
(2) tracking the speakers' emotions to make a more suitable reply~\cite{DBLP:conf/aaai/ZhouHZZL18}; 
(3) responding with more diverse sentences to avoid a boring user experience~\cite{DBLP:conf/naacl/LiGBGD16}; and 
(4) generating responses with personas to make the dialogue more consistent~\cite{DBLP:conf/ijcai/QianHZXZ18}.

\noindent\textbf{Personalized Chatbots.}
Building a personalized chatbot has attracted public interest in recent years. 
With a more steady personality, chatbots can have more consistent and informative dialogues with users.
Existing methods can be summarized as:
(1) Learning with user ID information~\cite{DBLP:conf/acl/LiGBSGD16,Rami_CCC_2016,Bak_VHUCM_2019,DBLP:conf/emnlp/ChanLYCHZY19}. 
These methods embed user IDs into vectors and use them to guide the dialogue. However, it is impractical to maintain a huge user ID embedding table for real-world large-scale datasets.
(2) Learning with explicit user profiles~\cite{DBLP:conf/ijcai/QianHZXZ18,Zhang_2018_dog,Alan_2019_PreGan, DBLP:conf/ijcai/SongZCWL19}. 
This group of methods aims at using descriptive information (such as persona sentences or attribute tables) to generate personalized responses, but such information is hard to collect, and it cannot keep track of user interest changes.
(3) Learning with implicit user profiles~\cite{DBLP:conf/sigir/madousigir21,DBLP:conf/naacl/ZhongD0QW22}. 
These methods extract personalized information automatically from a user's dialogue history to generate personalized responses. 

However, user dialogue history is limited and persona-sparse, 
and it is challenging to learn robust representations of the user's implicit profile.
Therefore, we aim to tackle this problem and enhance representations to generate personalized responses.

\section{Methodology} \label{model}
In this section, we first provide the problem statement and an overview of our model. 
Then, we elaborate the details of each component in the model. 

\subsection{Problem Statement and Overview}

\begin{figure}[t]
\centering
    \includegraphics[width=\linewidth]{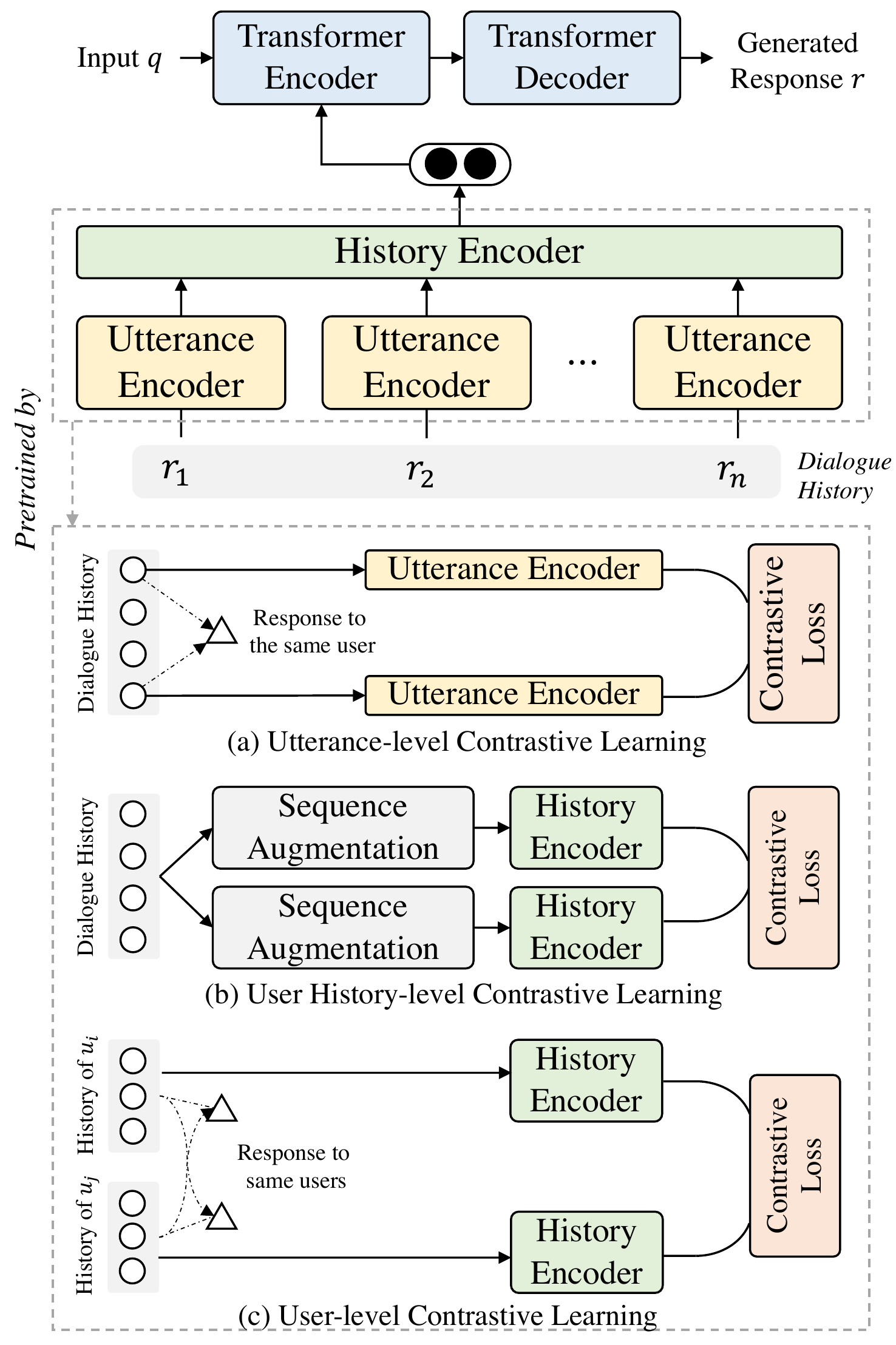}
    \caption{The proposed three pre-training tasks based on (1) Response Pairs, (2) Sequence Augmentation Pairs, and (3) User Pairs.}
\label{fig:model}
\end{figure}

Considering a set of users defined as $U = \{ u_1, u_2, \cdots, u_l \} $, for a specific user $u_i$, their dialogue history is denoted as 
$H^i=[(q^i_1, r^i_1, t^i_1), (q^i_2, r^i_2, t^i_2), \cdots, (q^i_n, r^i_n, t^i_n)]$, 
where $n$ represents the length of user dialogue history, and $t^i_k$ represents the response time of $r^i_k$.
Note that $q^i_j$ is a query issued by other users, while $r^i_j$ is the response given by the user $u_i$. We only use the historical responses that reflect the personal information of the user $u_i$, so $H^i$ is simplified as $[r^i_1, r^i_2, \cdots, r^i_n]$ henceforth.
With the above notations, our task is defined as generating a personalized response $r$ for the user $u_i$ to reply to a new query $q$ with the personalized information extracted from the user's dialogue history $H^i$.

The overview of our MCP model is shown in Figure~\ref{fig:model}. For user dialogue history, we design an utterance encoder to independently represent each utterance into vectors; and a history encoder to model the interactions among utterances and capture their sequential information. The obtained history representation (containing personal information) and the input query are fed into a Transformer-based Seq2Seq structure to generate a personalized response. 

As the main contribution of this work, we design three self-supervised pre-training tasks for an utterance encoder and a history encoder. By sampling contrastive samples at utterance\nobreakdash-, dialogue history\nobreakdash-, and user-level, the two encoders are trained to capture more personalized information (such as user interest) from user dialogue history and create more robust representations. We will elaborate the details of the three pre-training tasks and our model in the following sections.

\subsection{Utterance-level Contrastive Learning}
The first contrastive pre-training task is based on our observation on conversations between two people: the consecutive utterances within a short period are usually about one topic. 
Intuitively, the representation of two utterances under similar topics in the dialogue history should be closer than other utterances. Following this assumption, we extract contrastive samples from the dataset by the rule: the two utterances should be replied to the same user and their issued time should be close.
Formally, given a user $u$, we select two triplets $(q_i, r_i, t_i)$ and $(q_j, r_j, t_j)$. Both queries $q_i$ and $q_j$ are issued by the same user $u'$, and the response time interval satisfies $|t_i - t_j| \leq \tilde{t}$, where $\tilde{t}$ is a time threshold.
As a result, $r_i$ and $r_j$ construct a pair of contrastive samples. We then apply a contrastive learning objective to pull close their representations and push apart the representation of other utterances. 

Specifically, we use an utterance encoder to represent the responses $r_i$ and $r_j$ as:
\begin{align}
 \overline{\br}_i &= \operatorname{Mean}(\operatorname{Transformer}_{\text{u}}(r_i)),\notag \\
 \overline{\br}_j &= \operatorname{Mean}(\operatorname{Transformer}_{\text{u}}(r_j)),\notag
\end{align}
where $\operatorname{Mean}(\cdot)$ is the mean pooling operation.
The general contrastive learning objective aims at reducing the distance of positive pairs and increasing the distance of negative pairs~\cite{DBLP:journals/jmlr/GutmannH10}.
In our case, for a positive response representation pair $(\overline{\br}_i, \overline{\br}_j)$ in a mini-batch of $N$ pairs, we consider the response $\overline{\br}$ in the other $N-1$ pairs as negative set $R^-$. Hence, the loss function can be defined as:
\begin{align}
    \mathcal{L}_{\text{utt}} &= -\text{log}\frac{\phi(\overline{\br}_i, \overline{\br}_j)}{\phi(\overline{\br}_i, \overline{\br}_j)+
    \sum\limits_{\overline{\br}' \in R^-}{\phi(\overline{\br}_i, \overline{\br}')}},\notag \\  
    \phi(\overline{\br}_i, \overline{\br}_j) &=
    \text{exp}(\text{CosSim}(\overline{\br}_i, \overline{\br}_j)),\notag
\end{align}
where the function CosSim($\cdot$) denotes the cosine-similarity.
\subsection{History-level Contrastive Learning}

In personalized response generation, the main task is to capture the personalized information (\eg, the user's interests) from the dialogue history. To enhance the history encoder and obtain a more robust history sequence representation, we apply sequence augmentation to construct different views of the dialogue history. By comparing the original sequence and the augmented one, the history encoder is encouraged to highlight the information that best reflects the user’s personality. Specifically, we design the following four strategies:

(1) \textbf{Session masking.}
When chatting with others, a user often shows the same interest to different people. It inspires us that after masking all responses to a single person, the rest of the user's dialogue history can still represent the user's intrinsic interest. 
Based on this assumption, for each user, we select a series of her responses that respond to the same user and then mask them in the sequence of her dialogue history.
The altered sequence and the original sequence construct a positive pair.
With this task, we could improve the generalizability of the user profiling model.

(2) \textbf{Sequence random masking.}
Generally, a user's long-term interests are stable. 
This inspires us to randomly mask $k$\% sentences in sequence to prevent our model from overly focusing on several certain responses. 
The new sequence and the original one construct a positive pair.
The unmasking part of the dialogue sequence doesn't affect the user's long-term interests. Furthermore, the robustness of our model is enhanced with this task.

(3) \textbf{Sequence re-ordering.}
Although the order of user dialogue history has an effect on the user profile, 
simply re-ordering two responses from dialogue history should not affect the user's long-term and stable interests. 
Thus, we randomly choose pairs of responses that respond to different people and swap their positions in the sequence.
The augmented sequence and the original one construct a positive pair. 
The long-term and stable interests are highlighted with this task.

(4) \textbf{Short interval sequence masking.}
Inspired by research on session-based social recommendation~\cite{DBLP:conf/ksem/ZhangBWW20},
a pair of consecutive responses with tight time intervals imply more user interests.
Thus, we choose pairs from consecutive responses which satisfy the condition that time interval $t \leq \hat{t}$, where $\hat{t}$ denotes a time threshold.
By simply masking the latter response in the chosen pairs, the user profile should remain unchanged.

For a specific user dialogue history sequence $[r_1,\cdots,r_n]$, 
we augment it and obtain a new sequence as $[r'_1,\cdots,r'_n]$ following the aforementioned four strategies. These two sequences are treated as a positive pair and encoded by the history encoder as:
\begin{align}
 \textbf{U} &= \operatorname{Mean}(\operatorname{Transformer}_{\text{h}}([\overline{\textbf{r}}_1;\cdots; \overline{\textbf{r}}_n])),\notag \\
 \textbf{U}_{\text{aug}} &= \operatorname{Mean}(\operatorname{Transformer}_{\text{h}}([\overline{\textbf{r}}_1';\cdots; \overline{\textbf{r}}_n'])),\notag
\end{align}
where [;] is the concatenation operation. As the obtained vector containing the user's personalized information, we call it \textit{user profile} vector.
Thereafter, we get positive augmented pairs $(\textbf{U}, \textbf{U}_{\text{aug}})$.
Similarly, the negative set $S^-$ contains $N-1$ augmented sequences sampled from the mini-batch of $N$ size, and the loss is as follows:
\begin{align}
    \mathcal{L}_{\text{seq}} = -\text{log}\frac{\phi(\textbf{U}, \textbf{U}_{\text{aug}})}{\phi(\textbf{U}, \textbf{U}_{\text{aug}})+
    \sum\limits_{\textbf{U}' \in S^-}{\phi(\textbf{U}, \textbf{U}')}}.
\end{align}

\subsection{User-level Contrastive Learning}
Besides, we also notice that user relationships affect the similarity of user profiles.
For two users $u_i$ and $u_j$, the set of people they respond to can be respectively described as $U'_i$ and $U'_j$.
Following research towards a recommendation system on social chatting forum~\cite{DBLP:conf/sigir/Ma14},
the larger common group of users they chat with, the more common topics and interests they share with.
Thus, we only choose pairs which support $|U'_i \cap U'_j| \geq \hat{s}$, where $\hat{s}$ denotes a threshold.

After sampling among users, we get a positive pair of similar user profiles $(\textbf{U}, \textbf{U}_{\text{pos}})$,
and a negative set $\textbf{U}^-$ which consists of other user profiles in the mini-batch. The loss learned by this task is as follows:
\begin{align}
    \mathcal{L}_{\text{user}} = -\text{log}\frac{\phi(\textbf{U}, \textbf{U}_{\text{pos}})}{\phi(\textbf{U}, \textbf{U}_{\text{pos}})+
    \sum\limits_{\textbf{U}' \in \textbf{U}^-}{\phi(\textbf{U}, \textbf{U}')}}.
\end{align}

\subsection{Personalized Response Generation}
We apply an encoder-decoder architecture based on Transformer~\cite{Noam_transformer_2017} to generate personalized responses.
First, the Transformer encoder encodes both user profile $\textbf{U}$ and current query $p$ combined together.
Then, the decoding module decodes the outputs from the first step, added with a linear layer so as to project them into a space with a vocabulary-size dimension. The process can be defined as:
\begin{align}
    \mathbf{M} &= \operatorname{TransformerEnc}([\textbf{U}; p]), \\
    \hat{y} &= \operatorname{Linear}(\operatorname{TransformerDec}(\mathbf{M})),
\end{align}
where $\hat{y} \in \mathbb{R}^v$, and $v$ is the size of the vocabulary.
Later, $\hat{y}$ is normalized by a softmax layer and represents the generation probability:
\begin{align}
    \hat{y}_{\text{prob}} = \operatorname{softmax}(\hat{y}).
\end{align}

\subsection{Training and Optimization}
At the pre-training stage, our goal is to optimize the loss of the proposed tasks at three levels:
\begin{align}
    \mathcal{L}^p=    \mathcal{L}_{\text{utt}}+\mathcal{L}_{\text{seq}}+\mathcal{L}_{\text{user}}.
\end{align}
At the fine-tuning stage, the model is trained to maximize the generation probability of the ground-truth response $y$, which can be optimized by cross-entropy loss as:
\begin{align}
    \mathcal{L}^f= -y~\text{log}~\hat{y}_{\text{prob}}.
\end{align}

\begin{table*}[t!]
\caption{The results of automatic evaluations on Weibo and Reddit datasets. We categorize the baselines into four groups: (1) non-personalized; (2) using user-id; (3) using explicit user profiles; and (4) using dialogue history. The best results are in bold. ``$\dagger$'' indicates that our model achieves significant improvement over the state-of-the-art result in paired t-test with $p$-value $<$ 0.05.}
\label{tab: auto result}
\small
\centering
\begin{tabular}{llcccccccc}
\toprule
\multirow{2}{*}{Dataset}& \multirow{2}{*}{Model} & \multicolumn{3}{c}{Overlap-based Metric} & \multicolumn{3}{c}{Embedding Similarity} & \multicolumn{2}{c}{Personalization} \\  
\cmidrule(lr){3-5}\cmidrule(lr){6-8}\cmidrule(lr){9-10} 
& &  \multicolumn{1}{c}{BLEU-1} & \multicolumn{1}{c}{BLEU-2} & \multicolumn{1}{c}{ROUGE-L} & \multicolumn{1}{c}{Average} & \multicolumn{1}{c}{Extreme} & \multicolumn{1}{c}{Greedy} & \multicolumn{1}{c}{P-F1(\%)} & \multicolumn{1}{c}{P-Cover}\\ 
\midrule
\multirow{10}{*}{Weibo} & (1) Seq2SeqWA & 3.335 & 0.294 & 8.740 & 0.321 & 0.266 & 0.254 & 1.698 & 0.041\\

 & (1) MMI & 3.632 & 0.095 & 5.317 & 0.477 & 0.695 & 0.305 & 1.874 & 0.054\\
 
 & (2) Speaker & 4.997 & 0.113 & 7.993 & 0.492 & 0.712 & 0.311 & 2.119 & 0.082\\
 
 & (2) PersonaWAE & 3.503 & 0.155 & 11.305 & 0.513 & 0.724 & 0.307 & 5.108 & 0.093\\
 
 & (3) GPMN & 4.901 & 0.696 & 8.090 & 0.353 & 0.391 & 0.301 & 4.512 & 0.084\\
 
 & (3) PerCVAE & 5.115 & 0.299 & 7.952 & 0.469 & 0.659 & 0.299 & 3.817 & 0.086\\
 
 & (4) VHRED-P & 6.992 & 0.709 & 10.695 &  0.437 & 0.560 & 0.307 & 5.459 & 0.065\\
 
 & (4) ReCoSa-P & 7.266 & 0.844 & 11.469 & 0.419 & 0.510 & 0.312 & 5.717 & 0.061 \\
 
 & (4) DHAP & {9.324} & {0.894} & {14.122} & {0.523} & {0.747} & {0.313} & {7.013} & {0.144}\\

 & (4) {MCP} (Ours) & \textbf{9.714}$^\dagger$ & \textbf{0.952}$^\dagger$ & \textbf{14.601}$^\dagger$ & \textbf{0.542}$^\dagger$ & \textbf{0.783}$^\dagger$ &  \textbf{0.319}$^\dagger$ & \textbf{7.315}$^\dagger$ & \textbf{0.149}$^\dagger$ \\ 
\midrule
\multirow{10}{*}{Reddit} & (1) Seq2SeqWA & 1.819 & 0.023 & 4.069 & 0.545 & 0.554 & 0.472 & 0.516 & 0.029\\

 & (1) MMI & 2.065 & 0.011 & 3.784 & 0.543 & 0.607 & 0.454 & 0.828 & 0.038\\
 
 & (2) Speaker & 2.642 & 0.054 & 4.469 & 0.538 & 0.606 & 0.457 & 1.455 & 0.031\\
 
 & (2) PersonaWAE & 2.637 & 0.113 & 8.199 & 0.629 & 0.685 & 0.442 & 3.392 & 0.032\\
 
 & (3) GPMN & 2.686 & 0.376 & 4.776 & 0.406 & 0.331 & 0.358 & 3.026 & 0.037\\
 
 & (3) PerCVAE & 5.933 & 0.576 & 8.112 & 0.637 & 0.649 & 0.499 & 3.456 & 0.040\\
 
 & (4) VHRED-P & 5.802 & 0.648 & 8.345 & 0.558 & 0.635 & 0.472 & 3.772 & 0.047\\
 
 & (4) ReCoSa-P & 6.113 & 0.686 & 8.899 & 0.574 & 0.632 & 0.510 & 3.998 & 0.044\\
 
 & (4) DHAP & \textbf{6.858} & {0.737} & {11.720} & {0.709} & {0.721} & {0.539} & {4.639} & {0.111} \\
 & (4)  {MCP} (Ours) & {6.590} & \textbf{0.765}$^\dagger$ & \textbf{12.068}$^\dagger$ & \textbf{0.712} & \textbf{0.749}$^\dagger$ & \textbf{0.541} &  \textbf{4.919}$^\dagger$ & \textbf{0.117}$^\dagger$ \\
\bottomrule
\end{tabular}
\end{table*}

\section{Experiments} \label{settings}

\subsection{Datasets}
Following previous study~\cite{DBLP:conf/sigir/madousigir21}, we conduct experiments on two datasets from Chinese Weibo ~\cite{Li_pchatbot_2020} and English Reddit~\cite{DBLP:conf/acl/ZhangSGCBGGLD20}. Both datasets are collected from open-domain social media platforms, where users can post with various interests, and respond to other users in who they are interested. Each utterance in the dialogue has its own timestamp and user ID.
We consider that a training sample contains three parts: (1) a query; (2) a corresponding response; and (3) a dialogue history containing several responses issued before the current response. 

\noindent\textbf{Weibo Dataset}
Following~\cite{DBLP:conf/sigir/madousigir21},
we use a subset of the PChatbotW dataset~\cite{Li_pchatbot_2020} containing 300K users.
The dataset is collected from the Weibo online platform\footnote{\url{https://www.weibo.com}}.
We construct the corresponding query and response to a query-response pair.
Moreover, we compare the response timestamps and consider the replies before the current response as the dialogue history.
Also, we refer to the method introduced in~\cite{Li_pchatbot_2020}.
Specifically, we remove hashtags, URLs, emojis and swear words from sentences. 
Then, we remove sentences that include more than 100 words or less than 5 words or contain multi-languages.

\noindent\textbf{Reddit Dataset}
To measure model performance in different language, we also introduce the English dataset, which is collected from the English social platform Reddit\footnote{\url{https://www.reddit.com}}.
In terms of the specific tree structure of Reddit, we consider the pair of parent node and its child node as a query-response pair.
We use the same cleaning process applied to the Weibo dataset to handle the Reddit dataset. It comprises 315.34K users.

\subsection{Baseline Methods}
We evaluate our model with four groups of relevant and typical baseline methods:

(1) \textit{Non-personalized response generation models: }
\textbf{Seq2SeqWA}
~\cite{DBLP:journals/corr/BahdanauCB14} applies attention~\cite{Luong_atten_2015} module to the GRU-based Seq2Seq model. 
\textbf{MMI}~\cite{DBLP:conf/naacl/LiGBGD16} optimizes the maximum mutual information loss to improve the diversity of generated responses. 

(2) \textit{{Personalized models using user ID:}} \textbf{Speaker}~\cite{DBLP:conf/acl/LiGBSGD16} improves the Seq2SeqWA model by incorporating user ID embeddings into the input of the decoder. \textbf{PersonaWAE}~\cite{DBLP:conf/emnlp/ChanLYCHZY19} generates personalized responses with a Wasserstein autoencoder. The user ID embeddings are used to build a Gaussian mixture distribution for personalization.

(3) \textit{{Personalized models using explicit user profiles:}} \textbf{GPMN}~\cite{Zhang_2018_dog} designs a memory module to encode and store the persona profile to enhance the Seq2Seq model. \textbf{PerCVAE}~\cite{DBLP:conf/acl/ZhaoZE17} applies a conditional variational autoencoder to improve the diversity of response.

(4) \textit{{Personalized models using implicit user profiles:}} \textbf{VHRED-P}~\cite{DBLP:conf/aaai/SerbanSLCPCB17} is a multi-turn response generation method that models dependencies among user's dialogue context. We replace the dialogue context with the user's historical post-response pairs to achieve personalization. \textbf{ReCoSa-P}~\cite{DBLP:conf/acl/ZhangLPGC19} is also a multi-turn response generation model that measures the context with an attention mechanism. Similarly, we simulate the context through the historical post-response pairs. \textbf{DHAP}~\cite{DBLP:conf/sigir/madousigir21} constructs general user profile from user response history, and establishes a key-value memory network to build a dynamic query-aware user profile. Then, the model generates personalized responses with a personalized encoder. This is the state-of-the-art method in personalized response generation.

\subsection{Implementation Details}
For a user, we use the former 80\% dialogue history to construct the training set, the middle 10\% part for the validation set, and the last 10\% for the test set. 
For each response, we use the corresponding query as the current query and the previous responses as the dialogue history.
The mentioned thresholds $\tilde{t}$, $\hat{t}$, and $\hat{s}$ are set respectively to 25\% and 5\% quantile time of all consecutive responses on datasets, and 1. 
The history length is set as 20, and the hyper-parameter $k$ is set as 30.
We initialize the utterance encoder by the parameters of \texttt{bert-base-chinese}\footnote{\url{https://huggingface.co/bert-base-chinese}} and \texttt{bert-base-uncased}\footnote{\url{https://huggingface.co/bert-base-uncased}} checkpoint respectively.
The number of Transformer layers is 6. The hidden size and number of heads of the Transformer are set respectively to 768 and 8. We use beam search with a beam width of 12 to decode words.

\subsection{Evaluation Metrics}
\noindent\textbf{Automatic Evaluation}. We introduce three groups of automatic evaluation metrics to measure the performance of our model.

(1) \textbf{Overlap-based Metric}. We use BLEU-1/2~\cite{DBLP:conf/acl/PapineniRWZ02} and ROUGE-L~\cite{DBLP:conf/acl/LinO04} to measure the $n$-gram overlap between the generated response and ground-truth response. (2) \textbf{Embedding Similarity}. We employ embedding-based metrics~\cite{DBLP:conf/emnlp/ChanLYCHZY19} to measure the semantic similarity between the generated response and the ground-truth one. (3) \textbf{Personalization}. Following previous studies~\cite{DBLP:conf/sigir/madousigir21}, we employ Persona-F1~\cite{DBLP:conf/ijcai/LianXWPW19} to measure the unigram F1 between the generated responses and historical responses, and Persona Coverage~\cite{DBLP:conf/ijcai/SongZCWL19} to calculate IDF-weighted word overlapping score between the generated responses and dialogue response history.

\noindent\textbf{Human Evaluation}. Considering the diversity of utterances in the real world, a response different from the ground-truth may also be appropriate. 
Therefore, we randomly select 100 test triplets (\ie, query, response, and user dialogue history) and manually evaluate their quality in terms of three aspects: readability, informativeness, and personalization, which are suggested by~\citet{DBLP:conf/emnlp/ChanLYCHZY19}. 
The detailed evaluation criteria are given in Appendix~\ref{human evaluation}.

\subsection{Experimental Results} \label{results}
\paragraph{Automatic Evaluation}.
The metric-based evaluation results are shown in Table~\ref{tab: auto result}. We can observe that our MCP method achieves the best performance on both datasets in terms of most metrics. The improvement is statistically significant (t-test with $p$-value
$<0.05$), which demonstrates the effectiveness of applying self-supervised tasks to pre-train the encoder. We also have the following observations:

\begin{table}[t!]
\small
\centering
\caption{Human evaluation results on Weibo dataset.}
\label{tab:human}
\begin{tabular}{lccc}
\toprule
Model & Readabi. & Informat. & Persona. \\ \midrule
(1) Seq2SeqWA & 2.10 & 1.85 & 0.19  \\
(1) MMI & 2.06 & 1.88 & 0.23 \\
(2) PersonaWAE & 2.07 &  1.99  & 0.36  \\
(3) GPMN  & 2.12 & 1.92 & 0.35  \\
(3) PerCVAE & 2.04 &  2.01  &  {0.39}\\
(4) DHAP & 2.26 & 2.09 & 0.56\\
\midrule
(4) {MCP (Ours)}  &  \textbf{2.32} & \textbf{2.15}  & \textbf{0.59} \\ \midrule
Ground-Truth & 2.69 &  2.35  &  0.84 \\ \bottomrule
\end{tabular}%
\end{table}

(1) In general, the models using dialogue history perform better than those using explicit user profiles or user-id. This indicates that the user's dialogue history contains sufficient personalized information and is a more suitable way for building personalized chatbots. Furthermore, using any kind of personalized information can improve the quality of the response. This confirms the effectiveness and necessity of applying personalization to response generation. 
(2) Compared with DHAP, MCP achieves better performance on all three groups of metrics, indicating that our model is capable of generating more fluent and personalized responses. The improvement obtained by our effective method can demonstrate the superiority of the proposed self-supervised pre-training tasks.

\noindent\textbf{Human Evaluation}.
Table~\ref{tab:human} shows the result of human evaluation on Weibo dataset. The Fleiss Kappa is around 0.51, which shows a moderate agreement by the annotators. 
Specifically, MCP outperforms DHAP by 2.65\%/2.87\%/5.36\% respectively on the three aspects. All these results demonstrate the superiority of MCP in generating more fluent, informative, and personalized responses, and the potential of contrastive pre-training tasks on personalized response generation.

\begin{figure}
\centering
\includegraphics[width=.9\linewidth]{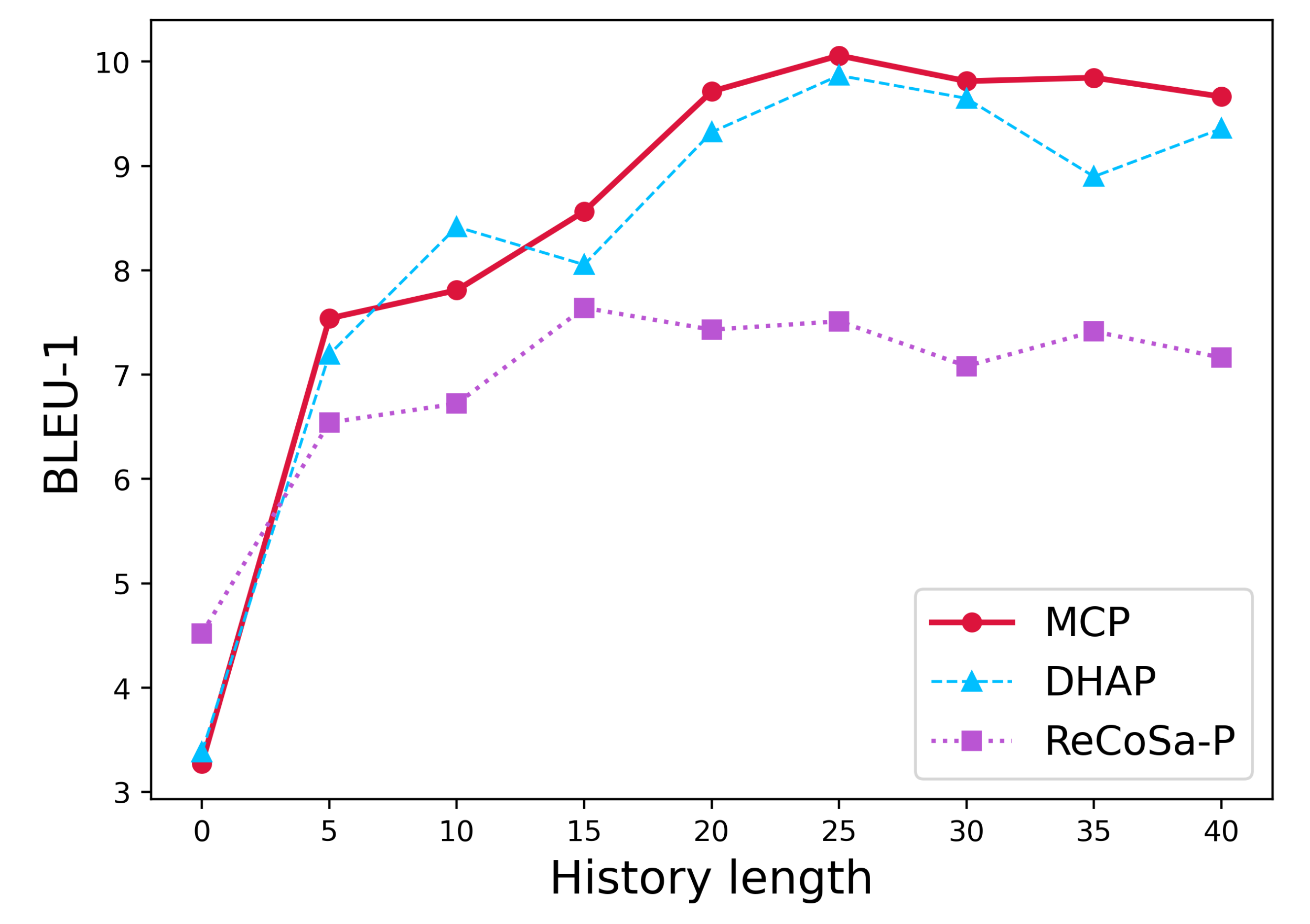}
\caption{Experiments with different lengths of user dialogue history on the Weibo dataset.}
\label{fig:HL}
\end{figure}

\begin{table}[t!]
\centering
\small
\caption{Ablation study results on Weibo dataset.}
\label{tab: ablation}
\begin{tabular}{lccc}
\toprule
Models & BLEU-1 & Avg. Emb. & P-F1 \\ \midrule
MCP (Full) & \textbf{9.714} &  \textbf{0.542} & \textbf{7.315} \\
\midrule
\quad \textit{w/o} Utter. Enc & 6.656 & 0.441 & 5.169 \\
\quad \textit{w/o} His. Enc & 7.548 & 0.489 & 5.742  \\
\midrule
\quad \textit{w/o} Pre. Utter. & 9.144  & 0.519 & 6.885\\ 
\quad \textit{w/o} Pre. His. & 7.893 & 0.497 & 5.969\\
\quad \textit{w/o} Pre. User & 9.438 & 0.534 & 7.146\\
\bottomrule
\end{tabular}%
\end{table}

\begin{figure*}[!t]
\centering
\includegraphics[width=.9\linewidth]{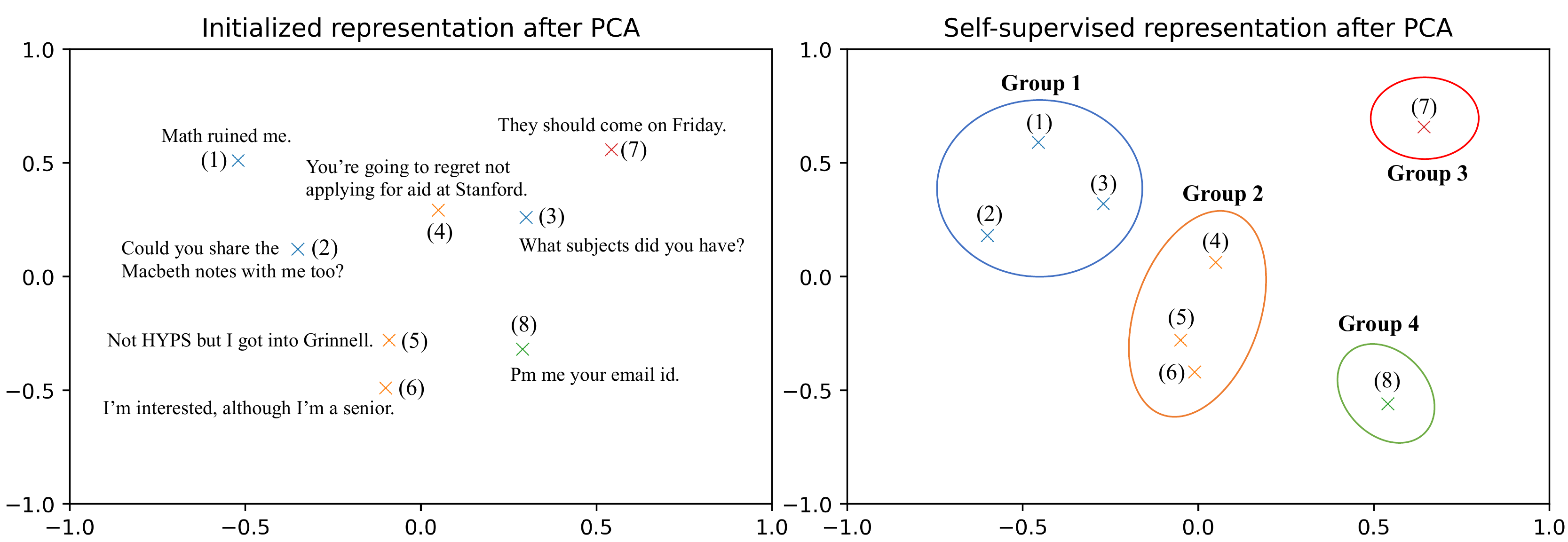}
\caption{The response representation distribution after PCA with scaling (before and after self-supervised pre-training) of user \#60 on the Reddit dataset.}
\label{fig:effect_sen}
\end{figure*}

\section{Further Analysis}
We further conduct a series of experiments to make an in-depth analysis of our model. The analyses are based on the results of the Weibo dataset (if not explicitly stated), and similar results can be observed from the Reddit dataset.

\begin{figure}[t]
\centering
\includegraphics[width=.9\linewidth]{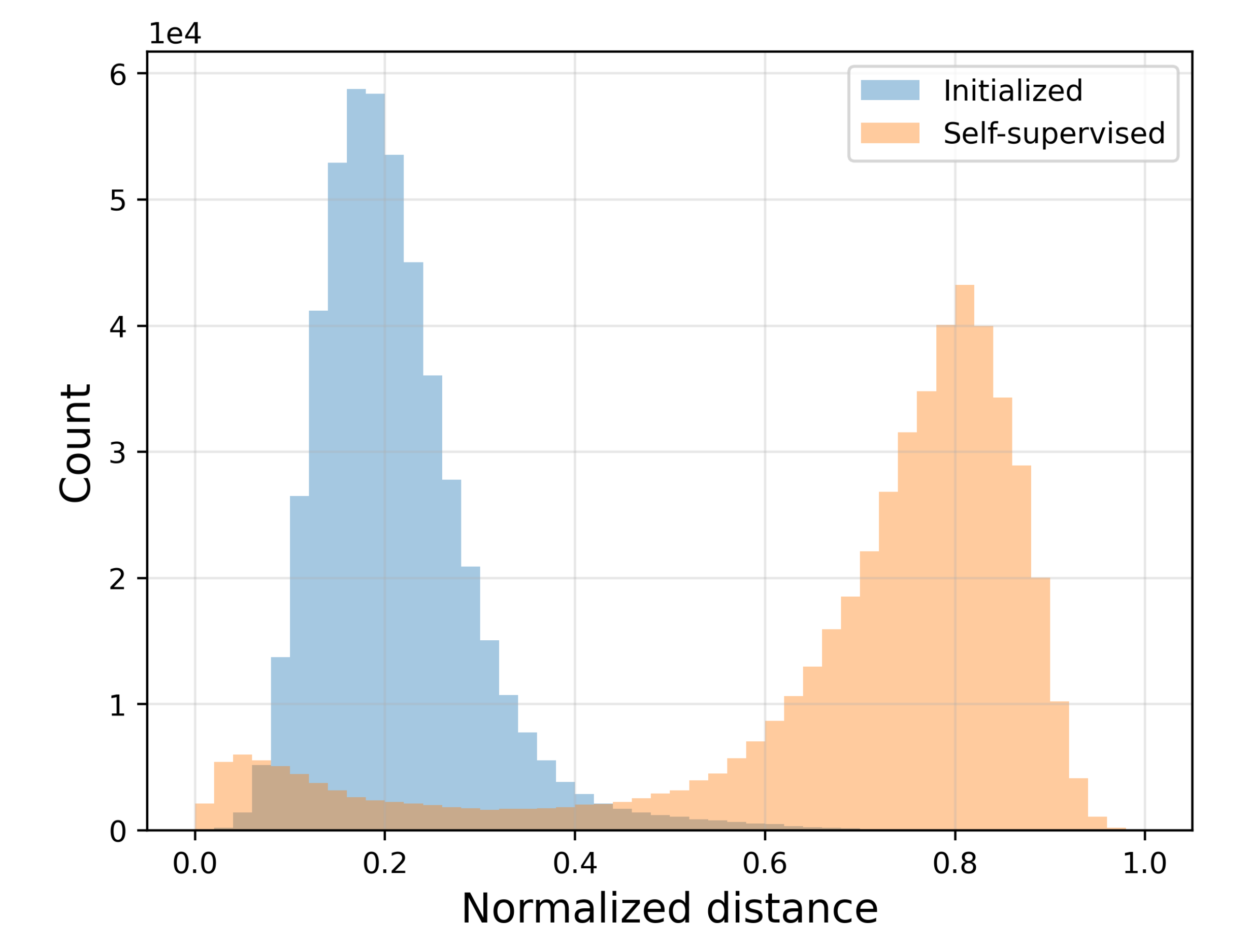}
\caption{Distribution of similarity between users on the Weibo dataset.}
\label{fig:effect_seq}
\end{figure}

\paragraph{Influence of History Length}
Since the personalized information is extracted from the dialogue history, the length of the history will influence the final performance. To explore such an influence, we conduct an experiment with MCP and two other strong baselines (DHAP and ReCoSa-P) by using various numbers of historical responses. The results are shown in Figure~\ref{fig:HL}. We have the following observations:
(1) In general, the performance of all three methods is increasing along with more historical responses used. This is consistent with our assumption that more dialogue history can provide more sufficient information. 
(2) MCP performs better with different lengths of dialogue history, indicating its robustness and generalizability. 
(3) When no dialogue history is provided, ReCoSa-P performs much better than both MCP and DHAP. The potential reason is that ReCoSa-P has a more complex structure to encode the user's input query and generate the response.

\paragraph{Ablation Study}
To investigate the effect of two encoders in MCP, we conduct an ablation study as:
(1) We remove the utterance encoder (\textit{w/o} Utter. Enc), and the utterance embedding is replaced by the mean pooling over all word embeddings.
(2) We remove the history encoder (\textit{w/o} His. Enc), and the user profile representation is replaced by the mean pooling over all utterance embeddings.
The results are shown on the upper side of Table~\ref{tab: ablation}. 
We can see that both encoders are important for our model since removing either of them leads to performance degradation. Concretely, the utterance encoder has a greater influence on the final performance. 
This is because the utterance encoder works at a lower level and is designed to capture fine-grained information from the utterance. If it is removed from the model, the poor utterance representation will also influence the history encoder. 
On the contrary, the history encoder aims at aggregating information from each historical utterance. 
It has relatively less influence when valuable information has already been refined into the utterance representation.

We further study the effectiveness of our proposed three pre-training tasks, and the results are shown in the lower side of Table~\ref{tab: ablation}. We denote the model without pre-training on utterance-/history-/user-level as \textit{w/o} Pre. Utter./His./User., respectively. It is clear to see that all three tasks are beneficial to our model. This demonstrates the great potential of contrastive learning on optimizing the representation of the utterance and dialogue history. 

\paragraph{Effect of Self-supervised Learning}
To investigate the effect of our proposed self-supervised learning, we measure the representation differences before and after pre-training. 

(1) \textit{Utterance Representation Changes.}
The purpose of the self-supervised pre-training task is to close the distance between responses sharing the same interests.
Thus, we randomly select a user from the Reddit dataset, extract historical responses, and map their embeddings to a two-dimensional space through PCA with scaling.
As shown in Figure~\ref{fig:effect_sen}, topic-related responses are marked with the same color and categorized into four groups.
In the initialized distribution, responses with the same color are dispersed, but they are obviously closer after self-supervised pre-training.
Moreover, the boundary between the four groups is wide and clear, which shows that the utterance encoder has the ability to sense different topic-related information in each response.

(2) \textit{History Representation Changes.}
The history encoder is designed to close the distance between representations of similar users.
The ability to distinguish users affects the personalization of generated responses.
Thus, we randomly select 1000 users from the Weibo dataset and calculate the distance between representations of every two users.
Figure~\ref{fig:effect_seq} shows that the initialized model cannot distinguish users well.
After training with self-supervised tasks, the differences between users are magnified, and the distance between similar users is reduced.
An interesting finding is that the group of dissimilar users is much larger than the group of similar ones.
This may benefit from our user-level contrastive learning task, which samples out similar users and closes their distance of representation.

\section{Conclusion} \label{conclusion}
In this paper, we proposed an MCP model for personalized response generation.
Different from previous personalized work,
first, we proposed a personalized responses generator that consists of an utterance encoder and a history encoder.
Next, we designed three self-supervised tasks from two levels to pre-train the two encoders.
Experimental results on two real-world datasets confirm the effectiveness of our model in generating informative and personalized responses.

\section*{Limitation} \label{limitation}
Though our method achieves promising results on two real-world datasets, there are still many limitations: 
(1) The architecture of our method is relatively simple compared with baseline methods. Other advanced structure for modeling the dialogue history may bring more improvement. 
(2) We have noticed that pre-trained language models have been applied to dialogue generation (\eg, DialoGPT~\cite{DBLP:conf/acl/ZhangSGCBGGLD20}). 
However, our method is currently based on a standard Transformer without pre-training on large-scale text datasets. 
Therefore, how to combine our proposed pre-training task with existing pre-trained language models is still under exploration.

\section*{Ethical Statement} \label{Ethical Statement}
The datasets used in this paper are available publicly online.
This paper does not contain any data collection or release, so there are no privacy issues.
The research will not pose any ethical issues.
We hired three well-educated and part-time annotators to conduct human evaluation with a pay of 100 CNY/hour during their evaluation.

\section*{Acknowledgments}
Zhicheng Dou is the corresponding author. This work was supported by the National Natural Science Foundation of China No. 61872370,  Beijing Outstanding Young Scientist Program NO. BJJWZYJH012019100020098,  the Fundamental Research Funds for the Central Universities, the Research Funds of Renmin University of China NO. 22XNKJ34, Public Computing Cloud, Renmin University of China, and Intelligent Social Governance Platform, Major Innovation \& Planning Interdisciplinary Platform for the ``Double-First Class'' Initiative, Renmin University of China. The work was partially done at Beijing Key Laboratory of Big Data Management and Analysis Methods, and Key Laboratory of Data Engineering and Knowledge Engineering, MOE.


\clearpage
\appendix

\section*{Appendix}

\section{Human Evaluation Criteria}\label{human evaluation}
We randomly select 100 test samples to conduct human evaluations.
The three well-educated annotators are shown with the generated responses, the corresponding query, and the user's historical query-response pairs.
Then the annotators will evaluate the generated responses from three aspects in a double-blind fashion.
The human evaluation criteria include: 
(1) Readability, which measures the grammatical correctness and fluency of the generated response;
(2) Informativeness, which measures if the generated response is informative or trivial;
and 
(3) Personalization, which measures whether the generated response shares some persona-related information with the dialogue history of the user.
The range of the first two factors is [1, 3], while the third is scored on a scale of [0, 1].
The specific scoring criteria for human annotation are shown in Table~\ref{tab:human criteria}.

\begin{table}[h!]
    \centering
    \small
    \begin{tabular}{l}
    \toprule
    \textbf{Readability}\\
    1: Hard to read or syntactically incorrect\\
    2. Good grammatical format\\
    3. Fluent and easy to understand\\
    \midrule
    \textbf{Informativeness} \\
    1: Meaningless responses\\
    2: Ambiguous or contain insufficient informative words\\
    3: Responses with clear meaning or purpose\\
    \midrule
    \textbf{Personalization} \\
    0: Contain no personal information in user history\\
    1. Reflect user's personal information in history\\
    \bottomrule
    \end{tabular}
    \caption{Specific scoring criteria of human annotation.}
    \label{tab:human criteria}
\end{table}
\end{document}